\documentclass{article}

\usepackage{color}
\usepackage{spconf,amsmath,graphicx}
\usepackage{epsfig}
\usepackage{graphicx}
\usepackage{amsmath}
\usepackage{amssymb}

\usepackage[colorlinks=true]{hyperref}

\begin{document}

\title{Multi-Scale Supervised Network for Human Pose Estimation}

\name{Lipeng Ke$^1$, Ming-Ching Chang$^2$, Honggang Qi$^1$, Siwei Lyu$^2$}
\address{
$^1$ University of Chinese Academy of Sciences, Beijing, China \\
$^2$ University at Albany, State University of New York, NY, USA 
}

%
%

%
\maketitle
%

\begin{abstract}
Human pose estimation is an important topic in computer vision with many applications including gesture and activity recognition. However, pose estimation from image is challenging due to appearance variations, occlusions, clutter background, and complex activities. To alleviate these problems, we develop a robust pose estimation method based on the recent deep conv-deconv modules with two improvements: (1) multi-scale supervision of body keypoints, and (2) a global regression to improve structural consistency of keypoints. We refine keypoint detection heatmaps using layer-wise multi-scale supervision to better capture local contexts. Pose inference via keypoint association is optimized globally using a regression network at the end. Our method can effectively disambiguate keypoint matches in close proximity including the mismatch of left-right body parts, and better infer occluded parts. Experimental results show that our method achieves competitive performance among state-of-the-art methods on the MPII and FLIC datasets.
\end{abstract}


\section{Introduction}
\label{sec:intro}
\vspace{-0.1cm}

\begin{figure}[t]
\centerline{
	\includegraphics[width=\linewidth]{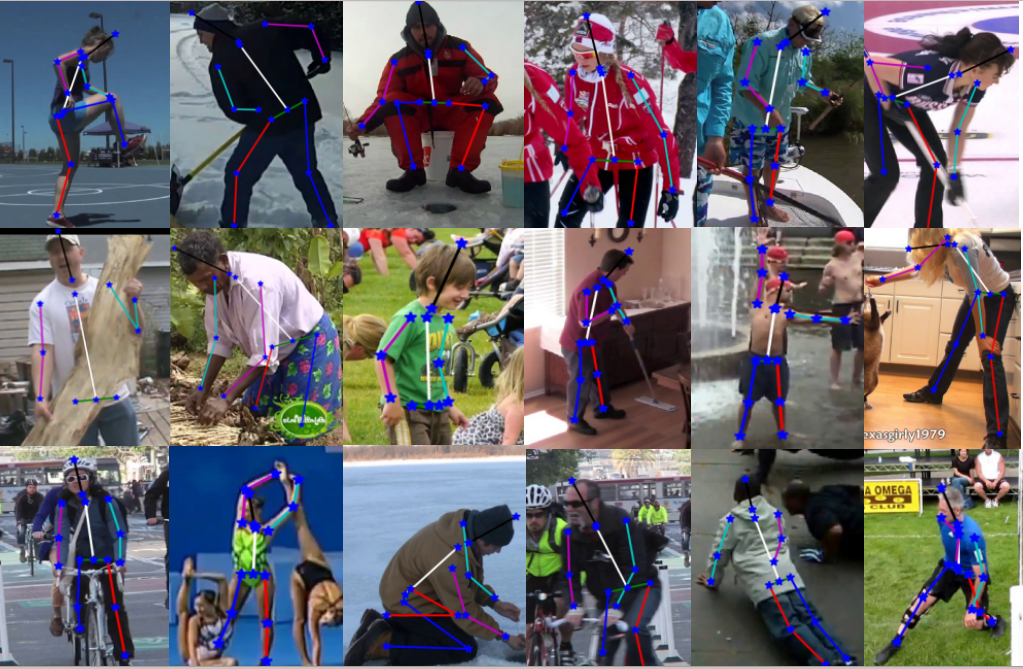}
\vspace{-0.2cm}
}
\caption{Examples of our human pose estimation on the MPII dataset. Our method can handle complex appearance, view variations, and diverse activities with heavy occlusions.
\vspace{-0.4cm}
}
\label{fig:mpii_example}
\end{figure}

Human pose estimation refers to the task of estimating body keypoint locations (wrists, elbows, knees, ankles, {\em etc.}) from images. This task can be very challenging due to the large variability of human body appearances, posture structures, the action being performed, viewing angles, occlusions, and complex backgrounds and lighting conditions; see Fig.~\ref{fig:mpii_example}.  Further sophistication of the inference is required when the cases extend to multi-person scenarios.

\begin{figure*}[t]
\centerline{
  \includegraphics[width=\linewidth]{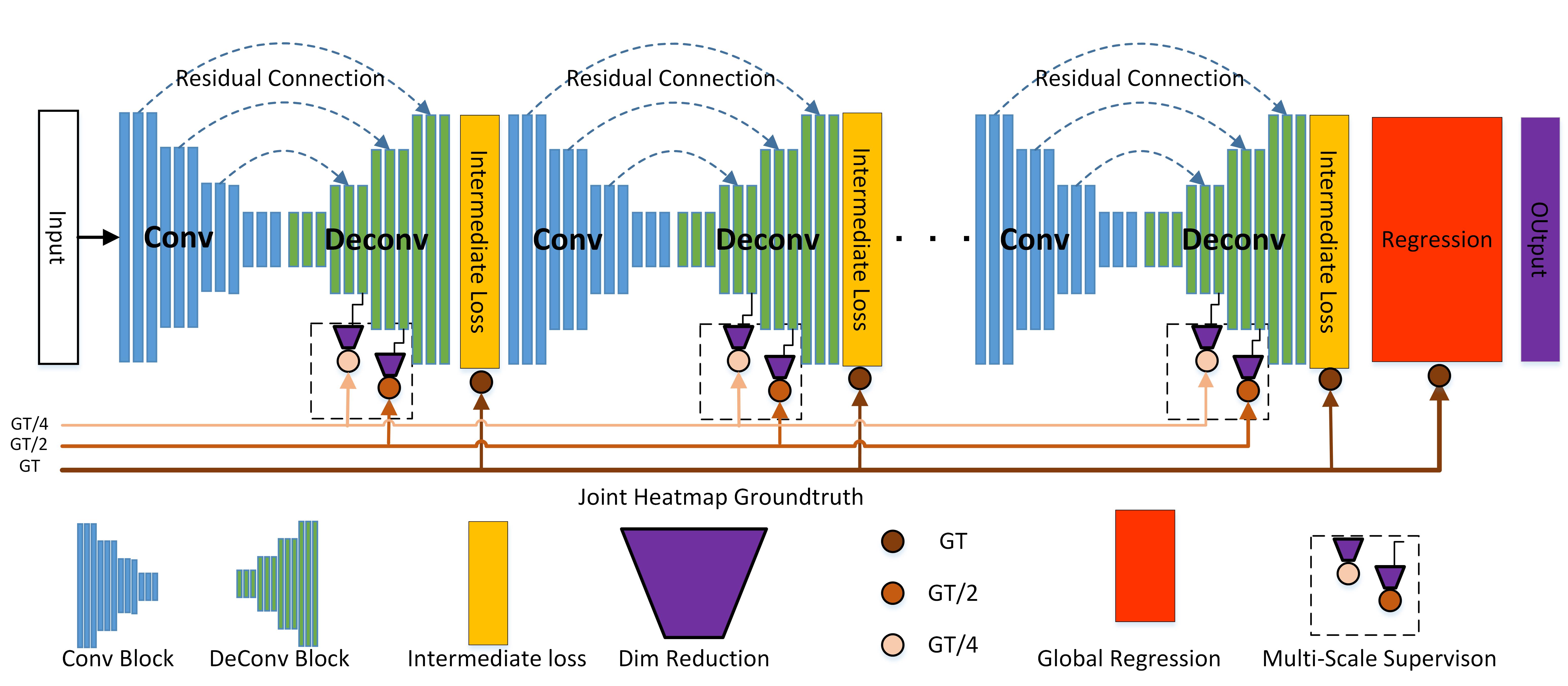}
  \vspace{-0.3cm}
}
\caption{
\label{fig:pipeline}
Our network model consists of four components --- {\em (i)} conv-deconv modules (blue and green, respectively), {\em (ii)} multi-scale supervisions (brown circles next to deconv layers), {\em (iii)} intermediate supervision layers (yellow), and {\em (iv)} global keypoint regression layers (red).
\vspace{-0.4cm}
}
\end{figure*}

Human pose estimation has been studied extensively \cite{liu2015a}. Classic methods including the use of histogram of oriented gradients (HOG) and deformable parts model (DPM) rely on hand-craft features \cite{Bourdev:Malik:Poselet:ICCV2009, Charles:BBC:Pose:BMVC2013, Cherian:MixingBodyPart:CVPR2014, Sapp:Taskar:MODEC:CVPR2013, chang_bmvc15}.
With the prosperity of Deep Neural Networks (DNN), Convolutional Neural Networks (CNN) have demonstrated remarkable performance boost in human pose estimation \cite{Toshev:DeepPose:CVPR2014, pfister2015flowing, Tompson:Unified:Pose:NIPS2014, Chu:Structured:Learning:Pose:CVPR2016, wei2016convolutional}. 
Tompson {\em et al.} \cite{tompson2014joint} adopted the {\em heatmap} representation of body keypoints to improve their localization during training. 
A Markov random field (MRF) inspired spatial model is used to estimate keypoint relationship. 
Chu {\em et al.} \cite{chu2016structured} propose a transform kernel method to learn local keypoint relationships, which is solved using a bi-directional tree. 

Recently, Wei {\em et al.} \cite{wei2016convolutional} use very deep sequential conv-deconv architecture with large receptive fields to directly perform pose matching on the heatmaps. They also enforce {\em intermediate supervision} between conv-deconv pairs to prevent gradient vanish.	
The hourglass module proposed by Newell {\em et al.} \cite{newell2016stacked} is an extension of Wei {\em et al.} with the addition of residual connections between the conv-deconv sub-modules. The {\em hourglass} module can effectively capture and combine features across scales. 
Chu {\em et al.} \cite{chu2017multi} adopt stacked hourglass networks to generate attention maps from features at multiple resolutions with various semantics. Yang {\em et al.} \cite{yang2017learning} design a Pyramid Residual Module (PRM) to enhance the deep CNN invariance across scales, by learning the convolutional filters on various feature scales.

State-of-the-art DNNs for pose estimation are still limited in the capability of modeling human body structural priors for effective keypoint matching.
Existing methods rely on a brute-force approach by increasing network depth to implicitly enrich the keypoint relationship modeling capability. A major weakness in this regard is the ambiguities arising from the occlusions, clutter backgrounds, or multiple body parts in the scene. In the MPII pose benchmark \cite{andriluka14cvpr}, many methods \cite{Chu:Structured:Learning:Pose:CVPR2016,wei2016convolutional,newell2016stacked,chu2017multi,yang2017learning} rely on repeating their pose estimation pipeline multiple times in various scales, in order to improve performance by a small margin using averaging of results. This indicates the lack of an effective solution to handle scale and structural priors in the modeling.



We propose a multi-scale supervised network model consisting of four components depicted in Fig.~\ref{fig:pipeline}.
Our main novelty is two-fold. First, we extend the intermediate supervision to explicitly cover multiple scales at the deconv layers during training. This improves the capability to extract more consistent and representative features across all scales. Our method can then effectively optimize feature representation across scales, because direct supervision is enforced at each scale during learning. 
Secondly, we use a regression network after the conv-dconv stacks to learn structural priors jointly from the keypoint feature maps from the conv-deconv stacks. This can effectively improve global pose estimation, when compared to existing methods \cite{wei2016convolutional,newell2016stacked,chu2017multi,cao2016realtime} which treat keypoint feature maps independently.

\vspace{-0.2cm}
\section{Multi-Scale Supervised Network Model}
\vspace{-0.3cm}



The proposed multi-scale supervised network is motivated by two key observations. First, in the existing works based on conv-deconv networks \cite{wei2016convolutional,newell2016stacked,chu2017multi,cao2016realtime}, accurate body keypoint correspondence depends largely on the consistency of the matching across multiple scales. 
This leads us to the design of multi-scale supervisions in training our network.
Secondly, since each body keypoint heatmap (corresponding to location likelihood) is estimated independently during the conv-deconv steps, structural relationship between individual keypoints are not modeled in the conv-deconv modules. To this end, we apply a global regression network at the end to model the keypoint relationship on top of the heatmaps. This improves the consistency of body structure in pose estimation in various scenarios: {\em (i)} to avoid left-right mismatches, {\em e.g.} matching a right arm to a left shoulder, {\em (ii)} to better handle occlusions, and {\em (iii)} to deal with multiple body parts and multiple people in the view.

%
%

\vspace{-0.2cm}
\subsection{Multi-Scale Supervision}
\vspace{-0.1cm}

We propose to enforce multiple supervision steps at individual deconv layers (shown in Fig.~\ref{fig:pipeline}) to learn richer multi-scale features for better keypoint localization.
As the depth of hourglass stacks increases, {\em gradient vanishing} becomes a critical issue during training.
{\em Intermediate supervision} \cite{wei2016convolutional} (yellow layers in Fig.~\ref{fig:pipeline}) between two conv-deconv stacks is a common practice, which by itself can address the gradient vanishing issue to some extent. 
However intermediate supervision at the original groundtruth scale dose not provide a consistent solution to cohesively supervise feature training across all conv-deconv scales. Our solution is then to apply supervisions to multiple scales of the deconv layers as shown in Fig.~\ref{fig:pipeline}. 


Our multi-scale supervision is an extension of the original intermediate supervision \cite{wei2016convolutional}. However our implementation to adopt the multi-scale design is different. 
Our multi-scale supervision is performed by calculating the residual in each scale regarding the down-sampled groundtruth heatmaps (denoted as GT/8, GT/4, GT/2) at each deconv layer in Fig.~\ref{fig:pipeline}. 
Specifically, to make consistent the feature map channels for the computation of keypoint groundtruth heatmap residuals at each scale, we use an 1-by-1 convolutional kernel (purple trapezoid in Fig.~\ref{fig:pipeline}) to convert the high-dimensional deconv feature maps into individual heatmap for each keypoint. This way, the dimension-reduced feature maps can be directly supervised against the respective scaled groundtruth using mean square error (MSE). 
We observe that our multi-scale supervision approach can improve the accuracy of keypoint heatmaps (with more focused distributions at keypoints) for use in the next deconv layer and  subsequent networks.
\footnote{If we remove the multi-scale intermediate supervision (GT/2 and GT/4 in Fig.~\ref{fig:pipeline}), and keep only the single-scale GT intermediate supervision (dark brown circle), as well as ignoring the regression network at the end, our network is reduced to an architecture similar to \cite{wei2016convolutional}.}

We describe our multi-scale intermediate loss terms {\em w.r.t.} the heatmaps of all keypoints as $L$2 loss in the following.
For the detection of $N$(=16) keypoints (head, neck, pelvis, thorax, shoulders, elbows, wrists, knees, ankles, and hips), $N$ heatmaps will be generated after each conv-deconv stack. 
The loss $L_i$ at the $i$-th scale compares the predicted heatmaps (of all keypoints) against the ground-truth heatmaps:
\begin{equation}
  L_i = \frac{1}{N}\sum_{n=1}^{N}\sum_{x,y}||P_n(x,y) - G_n(x,y)||_2 , 
\end{equation}
where $P_n(x,y)$ and $G_n(x,y)$ denote the predicted and the groundtruth heatmaps at the pixel location $(x,y)$ for the $n$-th keypoint, respectively. The total loss function is the summation across scales, $L=\sum_{i} L_{i}$, which is a combination of both the intermediate and multi-scale supervisions.



\vspace{-0.1cm}
\subsection{Global Keypoint Regression}
\vspace{-0.1cm}

%

\begin{figure}[t]
\centerline{
	\includegraphics[width=1\linewidth]{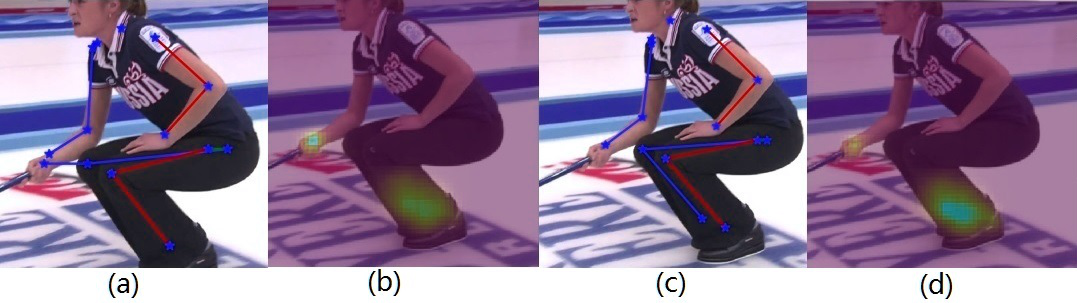}
\vspace{-0.3cm}
}
\caption{
Keypoint regression to disambiguate multiple peaks in the keypoint heatmaps. (a-b) shows an example of (a) the keypoint prediction and (b) the heatmaps from the conv-deconv module, which will be fed into the regression network. (c-d) shows (c) the output keypoint locations and (d) the heatmaps after regression. Observe that the heatmap peaks in (d) are more focused than in (b). 
\vspace{-0.2cm}
}
\label{fig:joint_regression}
\end{figure}

We use a {\em fully convolutional} regression network after the conv-deconv stacks to globally refine the multi-scale keypoint heatmaps to improve the pose structural consistency.
Our intuition is that the relative positions of arms and legs {\em w.r.t.} the head/torso represent useful action priors, which can be learned from the regression network by considering feature maps across all scales for pose refinement.
Our conv-deconv stacks extract heatmaps which are typically non-Gaussian according to the person's gesture/activity  (as shown in Fig.~\ref{fig:joint_regression}). The regression network then takes the multi-scale heatmaps as input, and match to the input image at respective scales. This way the regression network can effectively oversee the heatmaps across all scales for fine-tuning.

Specifically, heatmaps from the last conv-deconv stack together with their multi-scale heatmaps are concatenated and fed to the fully convolutional regression network.
Thus the pose structure is refined by the regressing feature map across all feature scales and body keypoints. This regression process can effectively refine keypoint locations in considering body structural priors. 
%
Fig.~\ref{fig:joint_regression}(c,d) shows an example of our multi-scale, across-keypoint fine-tuning with improved keypoint heatmaps and pose estimation accuracy.

\begin{table*}[t]
\caption{
Evaluation results on the MPII pose dataset (PCK$^h$=0.5)
\vspace{-0.3cm}
}
\begin{center}
\begin{tabular}{l||*{7}{c}|r}
  \hline
  & Head & Shoulder & Elbow & Wrist & Hip & Knee  & Ankle & Total \\ \hline
  Tompson {\em et al.} CVPR'15 \cite{tompson2015efficient}  &96.1 &91.9 &83.9 &77.8 &80.9 &72.3 &64.8 &82.0\\
  Belagiannis \& Zisserman FG'17 \cite{belagiannis2017recurrent}& 97.7  & 95.0  & 88.2  & 83.0  & 87.9  & 82.6 & 78.4 & 88.1 \\
  Insafutdinov {\em et al.} ECCV'16 \cite{insafutdinov2016deepercut}& 96.8  & 95.2  & 89.3  & 84.4  & 88.4  & 83.4 & 78.0 & 88.5 \\
  Wei {\em et al.} CVPR'16 \cite{wei2016convolutional}  &97.8 &95.0 &88.7 &84.0 &88.4 &82.8 &79.4 &88.5 \\ 
  Bulat \& Tzimiropoulos ECCV'16 \cite{bulat2016human} & {\bf 97.9} & 95.1  & 89.9  & 85.3  & {\bf 89.4}  & {\bf 85.7} & {\bf 81.7} & 89.7 \\ \hline
  Our model& {97.0} &{\bf 95.8} &{\bf 90.9} &{\bf 86.3}
           & {89.1} &{85.0} & {80.8} & {\bf 89.8} \\ \hline 
\end{tabular}
\vspace{-0.7cm}
\end{center}
\label{table:mpii-pckh5}
\end{table*}

\vspace{-0.1cm}
\section{Implementation and Experiments}
\label{sec:exp}
\vspace{-0.1cm} 

We train and test our model on two public datasets -- MPII (28K/12k train/test) \cite{andriluka14cvpr} and FLIC (5k/1k train/test) \cite{Sapp:Taskar:MODEC:CVPR2013} respectively.
Our stacked conv/deconv hourglass modules are trained 
on the respective datasets using the ADAM optimizer for 100 epochs, starting with initial learning rate 0.0005 with decay. 
%
Evaluations are described in three subsections.
$\S$\ref{sec:eval:accuracy} describes the accuracy evaluation on the two datasets. 
$\S$\ref{sec:eval:params} reports experiments regarding our network design and parameters, including the number of hourglasses and multi-scale supervision to investigate their effects regarding performance.
$\S$\ref{sec:eval:occlusion} evaluates how the multi-scale supervision can improve the handling of body part occlusions. 

\vspace{-0.3cm}
\subsection{Evaluation on Accuracy}
\label{sec:eval:accuracy}
\vspace{-0.1cm}

Evaluation is conducted using the standard Percentage of Correct Keypoints (PCK) metric \cite{tompson2015efficient}, which reports the percentage of keypoint detection falling within a normalized distance of the ground truth. For FLIC, PCK is set to the percentage of disparities between the detected keypoints {\em w.r.t.} the groundtruth after a normalization against a fraction of the torso size. For MPII, such disparities are normalized by a fraction of the head size, which is denoted as PCK$^h$. The PCK evaluation metric is defined as:
\begin{equation}
  \textrm{PCK}(\alpha) = \frac{1}{M}\frac{1}{N}\sum_{m=1}^{M}\sum_{n=1}^{N}I_{A} \left( \frac{||\textrm{Gt}_n-\textrm{Pred}_n||_2}{H} < \alpha \right),   
\end{equation}
where $M$ is the dataset size, and $N$ is the number of keypoints of a person. $I_A( \cdot )$ is an indicator function: $I_A$ is 1 if $( \cdot )$ is true, or 0 otherwise. $||\textrm{Gt}_n - \textrm{Pred}_n||_2$ is the Euclidean distance between the groundtruth and the prediction of the location of keypoint $n$. The normalization $H$ is half of the head size for PCK$^h$ and the torso size for PCK. Finally, $\alpha$ is the threshold to estimate if a keypoint is predicted correctly.

Table \ref{table:mpii-pckh5} summarizes the MPII performance evaluation. Observe that our method achieves state-of-the-art results across all keypoints (top 1 or 2, except the head) on the MPII dataset. 
Table \ref{table:flic-pck5} summarizes the FLIC results, where our PCK reaches 99.2\% for the elbow, and 97.3\% for the wrist. 
Our method performs better on shoulders, elbows, wrists that are in general harder to detect. This is due to improvements in our multi-scale feature supervision and global joint regression.


\begin{table}[t]
\caption{
\vspace{-0.5cm}
Results on the FLIC dataset (PCK=0.2)
\vspace{-0.3cm}
}
\begin{center}
  \begin{tabular}{l||*{7}{c}|r}
    \hline
    & Elbow & Wrist\\ \hline
    Tompson {\em et al.} CVPR'15 \cite{Tompson:Unified:Pose:NIPS2014} &93.1 &92.4 \\
    
    Wei {\em et al.} CVPR'16 \cite{wei2016convolutional}  &97.8 &95.0 \\ \hline
    Our model& \textbf{99.2} &\textbf{97.3} \\ 
    \hline 
  \end{tabular}
\vspace{-0.5cm}
\end{center}
\label{table:flic-pck5}
\end{table}

\begin{figure}[t]
\centerline{
\vspace{-0.0cm}
	\includegraphics[width=\linewidth]{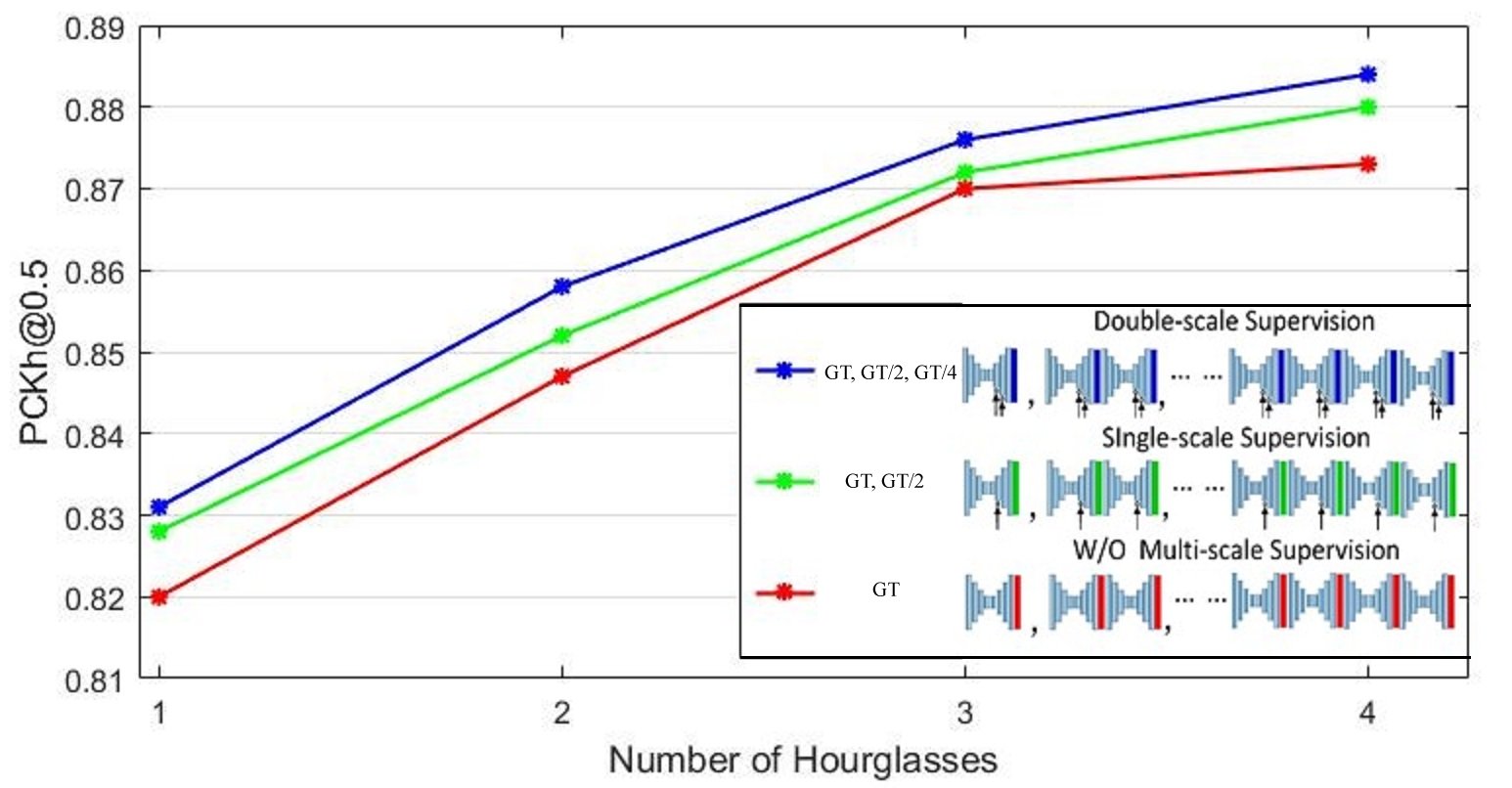}
\vspace{-0.3cm}
}
\caption{
Performance comparisons on the number of multi-scale supervision and network depth. Observe that increasing network depth (number of conv-deconv stacks) results in significant performance boost, since deeper network can extract better features for keypoint detection. Performance also increases with the number of multi-scale supervisions. 
\vspace{-0.3cm}
}
\label{fig:net-design}
\end{figure}

\vspace{-0.2cm}
\subsection{Evaluation on Network Parameters}
\label{sec:eval:params}
\vspace{-0.1cm}

We evaluate the components of the multi-scale network in two aspects on the MPII validation set, as shown in Fig.~\ref{fig:net-design}: (1) the number of conv-deconv stacks used in the network, and (2) the number of scales with intermediate supervisions: {\em (i)} groundtruth scale-only (GT, i.e. the original intermediate supervision as in \cite{wei2016convolutional} (red line), {\em (ii)} GT and GT/2 (green line), {\em (iii)} GT, GT/2, and GT/4 (blue line).

For pose estimation, a deeper network can mostly outperform a shallow one. However the network depth is limited by the available computational resource, especially the GPU memory used during training. State-of-the-art works \cite{wei2016convolutional,newell2016stacked,chu2017multi} use 4 GTX Titan X GPUs to run 8 conv-deconv stacks and 256 feature channels in the conv layers. In this paper, we use 4 conv-deconv stacks and 64 feature channels, in order to fit the model on a single GTX 1080 GPU. Our training resource is only $1/8$ of the state-of-the-art works.  

Fig.~\ref{fig:net-design} shows that the increasing use of conv-deconv stacks can consistently improve performance, which is expectable. It also shows that the increasing use of number of multi-scale supervision can consistently improve performance. 



\vspace{-0.3cm}
\subsection{Evaluation on Occlusion Handling}
\label{sec:eval:occlusion} 
\vspace{-0.1cm}




Occlusion is a common challenge for human pose estimation. 
We evaluate our method on a subset of MPII test set with available occluded keypoint labels. We focus on occluded keypoints which are connected to and can be inferred from other visible body parts, {\em e.g.} a hidden elbow can be recovered from visible shoulder and wrist locations. This experiment can evaluate how the proposed structural regression network performs hand-in-hand with multi-scale feature supervision for occlusion recovery. 
We obtain 86.7\% for PCK$^h$=0.5 with GT, GT/2, GT/4 multi-scale supervision.
In comparison, the score is 84.3\% without multi-scale supervision.


\begin{figure}[t]
\centerline{
	\includegraphics[width=\linewidth]{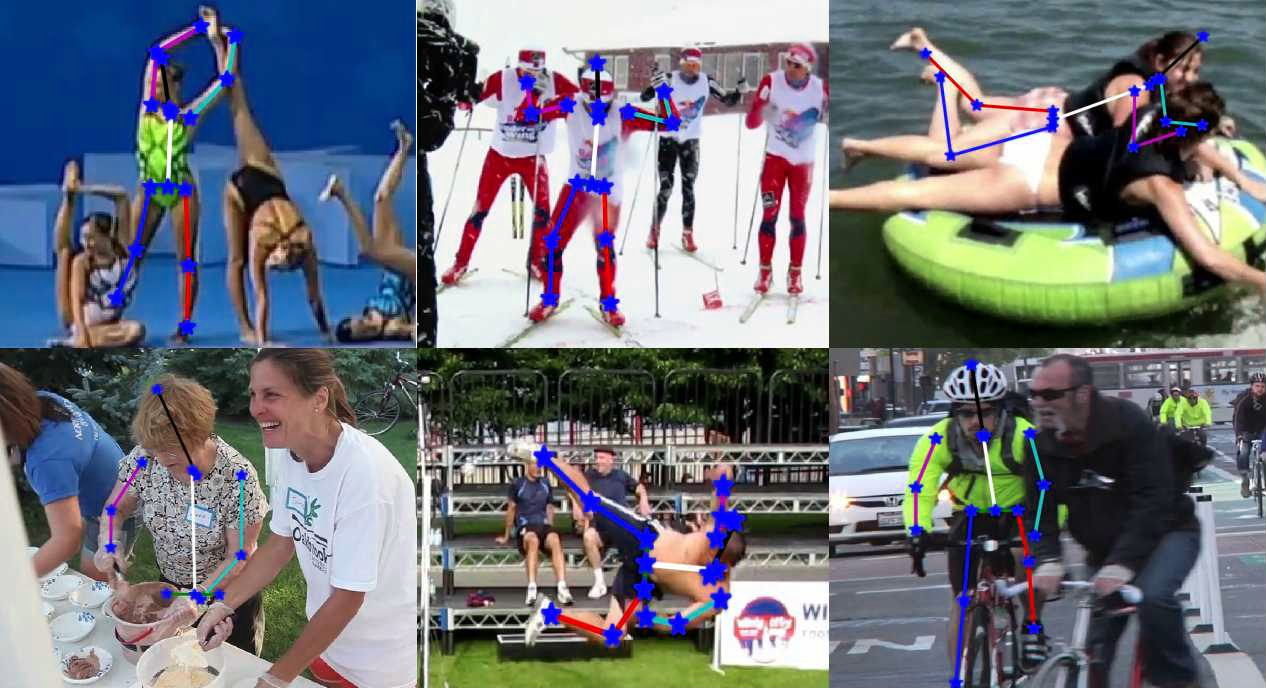}
\vspace{-0.3cm}
}
\caption{
Pose detection results on selected challenging samples from the MPII test set. These scenarios contain cluttered background, heavy occlusions, and activities involving multiple people near the subject of interest.
\vspace{-0.3cm}
}
\label{fig:hard_sampels_visual}
\end{figure}

 Fig.~\ref{fig:hard_sampels_visual} show our results on a few challenge cases in MPII test set involving multiple persons and complex background/occlusions. Observe that the proposed method can produce plausible results.

\vspace{-0.2cm}
\section{Conclusion}  
\vspace{-0.2cm}

We present an improved network with multi-scale supervision and structural keypoints regression for human pose estimation.
We show that both improvements can consistently increase performance when comparing with state-of-the-art methods.
Our method can effectively handle challenge cases including part occlusions, complex background and activities.

{\bf Future work} includes the use of deeper network stacks on multiple GPUs aiming for multi-person scenarios.





\bibliographystyle{IEEEbib}
\bibliography{refs}

\end{document}